# A Lexicon for Studying Radicalization in Incel Communities

Emily Klein and Jennifer Golbeck

## ABSTRACT

Incels are an extremist online community of men who believe in an ideology rooted in misogyny, racism, the glorification of violence, and dehumanization. In their online forums, they use an extensive, evolving cryptolect – a set of ingroup terms that have meaning within the group, reflect the ideology, demonstrate membership in the community, and are difficult for outsiders to understand. This paper presents a lexicon with terms and definitions for common incel root words, prefixes, and affixes. The lexicon is text-based for use in automated analysis and is derived via a Qualitative Content Analysis of the most frequent incel words, their structure, and their meaning on five of the most active incel communities from 2016 to 2023. This lexicon will support future work examining radicalization and deradicalization/disengagement within the community.

## INTRODUCTION

Incels (involuntary celibates) are an extremist community of misogynistic men that gather primarily in online forums. Although the incel community began as a place of commiseration and discussion among men and women who were struggling romantically, it morphed over time into a hotbed of misogyny, hegemonic masculinity, racism, self-hatred, and extremism (Ging 2017). Incels believe society has unfairly shunned them. They feel entitled to sex, fetishize very young women, believe that rape is justified, and that non-incels ("normies") should be punished for making them suffer. Fantasies of committing violence are common, and forums for incels have been repeatedly banned from major platforms because they encourage acts of violence, especially against women.

Because the community congregates almost entirely online, there is a wealth of data that supports analysis of topics like the language of extremism, radicalization, the social structure of extremist groups, community dynamics, and when violent fantasy tips over into violent action. As a first step toward analyzing the incel community more deeply, this paper presents an incel lexicon developed via a Qualitative Content Analysis of a comprehensive database of incel forums. The lexicon and this paper contain offensive content.

The goal of this work is to present a lexicon that is as complete as possible in terms of archetypal words and definitions to serve two main purposes. First, it provides background and context to lower the barrier to entry for researchers interested in analyzing incels. Second, it provides a basis for quantitatively analyzing the prevalence of incel-specific language in online spaces. That analysis may consist of tracking the frequency of this type of language over time for specific users, identifying its creep into non-incel spaces, or generating features for machine learning approaches to incel classification.

## BACKGROUND AND RELATED WORK

In recent years, increasingly lethal acts of violent extremism have been committed by incels, such as the 2018 Toronto van attack in which self-declared incel Alek Minassian drove a van into pedestrians, killing 10 and injuring 16. Minassian took inspiration from the 2014 Isla Vista, California attack where incel Elliot Rodgers killed 6 people and injured 14 others by gunshot, stabbing, and vehicle ramming. Rodgers is glorified by the incel community, which lauds him as "The Supreme Gentleman" or "Saint Elliot," regularly posting memes of his face photoshopped onto Christian icons.

Incels overwhelmingly voice support for acts of misogynistic violent extremism committed by members of their group and violence more generally (O'Donnell and Shor 2022). Waśniewska found that violence was a common metaphorical device, where sex with women is likened to war and conquest in which women are targets to be slayed (2020). Hoffman, Ware, and Shapiro argue that the core ideology of incels, combined with their enthusiastic support for violence and intent to have far-reaching societal effects should lead policymakers and scholars to consider incel violence as a new variety of terrorism with a distinct hate crime dimension (2020). They also note that sex and sexual frustration is a major motivating factor in radicalizing both incels and Islamic State members. One of the intended next analyses with this work will be to use terms from our lexicon to study the radicalization and deradicalization process. The boon of an incel-specific lexicon is that we have an inventory of terminology to serve as markers of language adoption and disuse over time.

The internet is a major decision-shaper when it comes to radicalization (Hassan et al. 2018). In a systematic review of empirical evidence on how the internet may or may not contribute to violent extremism, Hassan and colleagues found that exposure to radical violent material online, and actively seeking radical violent material seem to be associated with engaging in real-life political violence among white supremacist, neo-Nazi, and radical Islamist groups (2020). The frequency and dynamics of use of the terms in our lexicon may help quantify engagement with the community over time, as well as the level of active engagement versus passive engagement which may help identify individuals most likely to carry out misogynistic violent extremist attacks.

There is an extensive body of work in psychology on the social dynamics that lead people to adopt extreme beliefs. We know when people are drawn into fringe movements, cults, and extreme ideologies, many psychological factors are at play. Humans have a need to belong (Baumeister & Leary, 1995; Walton et al., 2012), and interpersonal connections to people in ideologically extreme groups can motivate people to join (Stark & Bainbridge, 1980). Groups with inspiring ideals that strike a chord with an outsider can provide a gateway to joining a group, and commitment to those ideals can override compunctions regarding extreme behavior the person might not otherwise have been willing to engage in (McAdam, 1986; Thrash & Elliott, 2003; Swank & Fahs, 2011). When a person's social identity is tied to an ideal, this can build commitment to a group that connects itself with that ideal (Ysseldyk, 2010). Once a sense of belonging or commitment to a group is created, it can feed deeper commitment and the growth of in-group bias (Efferson et al., 2008)

When engaging with an extremist group, social dynamics are key drivers of the radicalization process. In instances where in-group behaviors strongly diverges from the norm, newer members look to others for social proof, guidance as to what behavior will be rewarded within the new community (Cialdini et al., 1999). Rather than simply establishing an in-group culture, social proof can drive people to extreme behavior, including acts as extreme as suicide (Ji et al., 2014). Finally, when belonging to an extremist group comes with incentives (material or psychological) that are tied to performing certain beliefs, people's genuine beliefs will often shift to match what they are required to express (Solley & Santos, 1958; Festinger & Carlsmith, 1959).

In the incel space, this raises questions about the connection between the social community of the forums, dehumanizing and hateful language, and radicalization. Incels are generally socially isolated (Sparks et al., 2023), so the online community fills an important social and emotional need. The community also rewards – and often requires – participating in their language use and expressing extremist ideas. For new members, seeing that the community rewards others when they post extreme, violent, hateful content serves as social proof. It is possible that men who are not necessarily committed to the incel ideology will follow that guidance and perform extremism verbally to receive the social support of the community. If so, that may lead them to eventually genuinely adopt those beliefs; essentially, they radicalize by saying what is necessary to maintain these "friendships" online.

Incels' extensive insider vocabulary would be termed a "cryptolect". As an example, consider this representative message from the incels.is message board: "Finally, a cumskin acknowledges that JBW is a thing though the guy in the video is most definitely betabuxxing or cuckmaxxing bc toilets don't voluntarily go for n*****s." [censored by authors]. A non-jargon translation reads: "Finally, a white guy acknowledges that white men are favored in dating although the guy in the video is most definitely providing money or being subservient because women don't voluntarily go for black people."

Much of the language is dehumanizing to women, but it also serves as shorthand for common aspects of incel ideals and ideology, including a racist and generally dehumanizing worldview. Use of it marks membership in the community and effectively limits outsiders from dropping in and understanding their discussions. As a result, use of this language can be a proxy marker for incel radicalization; the more people use these terms, the more they are signaling both their membership in the community and adoption of their philosophy.

Incels often coin new words. Bogetić (2023) focused on this specific type of language in incel communities, noting that common characteristics were developing words to refer to people by their group membership. Members of groups were often referred to with terms derived from proper names (e.g. referring to specific types of women as *Stacy*), food (e.g. using *curry* to refer to people of Indian descent), or skin color.

Such objectifying language has clear ties to racist language, and incels themselves often explicitly assert that they are creating words with the intent of being derogatory. Waśniewska (2020) has also investigated the language of incels, with a focus on the origins of incel language (including pick-up artist forums and popular psychology) and on the symbolism and metaphor in their terminology. Her work concludes that the incel worldview, expressed through language, is hierarchical, pessimistic, and highly polarized. This polarization between people who are in the group (fellow incels) and people who are outside the group (normies) binds the community together through holding a shared, stigmatized worldview.

There are some informal guides to incel terminology. Gothard (2020, 2021) used a computational linguistics approach to identify words whose frequency in incel forums was highly divergent from a background dataset. Her work highlights many core incel terms and illustrates their use over time, but the goal of that work was not to create a comprehensive lexicon or guide to incel terms. Baele et al. (2023) also looked at the user of language over time in incel groups. They examined the valence of incel language with attention to violence and found that while language varies across forums, levels of violent language have increased steadily since 2017. Jaki and colleagues (2019) profiled users of the forum incels.me, broadly classifying language on the forum in terms of misogyny, racism, and homophobia. Other resources describe the ideology of the incel community very thoroughly, but may be limited in the depth and breadth necessary for outsiders to interpret examples like the one provided above. An incel lexicon is an important resource for scholars and policymakers studying the community, as well as for the broader public as incel terminology and ideology enter mainstream discourse and change over time (Solea and Sugiura 2023). To address this need, we provide a list of terms and their definitions, highlighting the combinatorial aspect of prefixes and suffixes that are generated by incels' creative blending of words.

## MATERIALS AND METHODS

To develop this lexicon, we used the Radicalization and Deradicalization in Online Communities dataset available at https://doi.org/10.7910/DVN/MS9ODP under a CC BY-SA 4.0 license (Golbeck, 2023). This dataset contains the full archive of five active incel communities: incels.is, the most popular active incel forum; the Saidit.net incel forum; the now-defunct Reddit /r/Incels and /r/Braincels forums, which were the core of the Reddit incel community before they were banned; and /r/IncelExit, a community for men leaving the incel community. The dataset contains 9.2 million posts from 122,079 users that span from 2016 to 2023.

We identified the most frequently occurring incel-specific words and affixes in the dataset using the log ratio to determine how much more prevalent a word/affix was compared to other words/affixes in the dataset for the most common 1000 words. From this log ratio list of most frequent words and affixes, we discarded common words such as *complete*, *problem*, *much*, *boyfriend*, *anime*, etc. that are not exclusive to the incel community. This left us with 64 words and affixes specific to the incel community. We carried out a qualitative content analysis, coding each term as belonging to one or more of the categories "dehumanizing," "racist," or "misogynistic." Three instances of ambiguity were resolved through discussion and close reading of the data, and no disagreements in coding occurred ($\kappa = 1$). We then assigned each term a definition based on the term's usage in context of the dataset. Here again, no disagreements between the authors occurred. The final lexicon is a list of 64 terms and their definitions, each coded as being dehumanizing, racist, and/or misogynistic. Unlike other existing lists of incel terminology, this list (1) is derived quantitatively from the most well-known and active incel communities over time, (2) includes affixes as essential building blocks of word formation which serves an important function in small communities, and (3) categorizes terms into three concept-driven thematic categories.

## RESULTS

The Incel Lexicon, with definitions, is included as Appendix A. We have also made a list with the roots, prefixes, and affixes, plus some of the most common blended words, available at <ANONYMIZED LINK> for use in computational linguistic techniques.

In our analysis of the incel lexicon, we focused on words that were mostly exclusive to incel communities. We did not include words that were common in other online spaces, even if they are widely used by incels. For example, the term "redpilled" derives from the movie *The Matrix* and generally refers to someone who has accepted harsh reality. It is used in the manosphere (the broad collection of online communities promoting various degrees of misogyny) to refer to men who believe society largely discriminates against men rather than women, but it is not incel-specific. Incels generally use a lot of highly-online language, memes, and Reddit slang, but since those terms are used in other communities, we did not consider them.

Word blends appear to be a hallmark of the incel lexicon. Blends are combinations of parts of words that have an irregular, unpredictable relationship between form and meaning, as in *frenemy* (friend + enemy), *affluenza* (affluent + influenza), or *bit* (binary unit). Word blends are challenging to study in comparison to word compounds, which are combinations of complete words or morphemes. Compounds have a regular, predictable relationship between form and meaning like *swimming pool* (a pool for swimming) or *toothpaste* (a paste for brushing one's teeth). Blends do not conform to regular word formation processes, but are nonetheless incredibly productive and creative, greatly enriching vocabulary to reflect new concepts (Hamawand 2011, Lahlou and Ho-Abdullah 2021). A unique feature of blends is that they can generate new affixes, e.g. *-gate* (as in Watergate, Pizzagate), *-gasm* (as in orgasm, nerdgasm, foodgasm), or in the case of an incel lexicon, affixes such as *curry-*, *chad-*, *-maxx*, *-fuel*, and *-mogg*.

Affixes introduce a semantic change to the stem word. For example, *statusmaxxing* means to maximize one's social status (and implicitly, sexual market value and desirability to women), and *looksmaxxing* means to maximize one's looks for the same goal. The suffix *-maxx* denotes maximization of the stem (status, looks, money, career, etc.) to improve one's success with women. Similarly, *chad-* is combined with the Sikh name suffix *-preet*, the Arabic name *Saddam*, and the given name *Tyrone* to refer to Black Americans. Thus, *chadpreet*, *chaddam*, and *chadrone* are "chad" versions of their namesake ethnicity. Stereotypical names, foods, and geographic characteristics belonging to, consumed by, or home to members of an ethnic or national group are often used as stand-ins for that group, e.g. *chadriguez* (a chad of Hispanic descent), *ricecel* (an incel of East or Southeast Asian descent), *currycel* (an incel of South Asian descent), *burgercel* (an American incel), and *sandcel* (an incel of Arabic descent). Waśniewska describes the metonymic use of food to represent race as reflective of the openly racist incel worldview (2020).

The suffix *-cel* denotes an involuntary celibate, where the stem denotes a characteristic of the incel (an aspect of their identity or selves) that is perceived to have contributed to their lack of success with women (e.g. *wristcel* for having thin wrists). Stems may also denote an activity the incel pursues to cope with being unsuccessful with women (e.g. *gymcel* for going to the gym to work out). The suffix *-mog* denotes domination by another, where in most cases the stem indicates the area in which someone dominates, as in *heightmog* (to dominate another man in height), or *skullmog* (to dominate another man in shape or size of the skull). Sometimes the stem indicates a specific person who dominates others (e.g. *dolphmogged* meaning dominated by Swedish actor Dolph Lundgren, or *deppmogged* meaning dominated by American actor Johnny Depp). Typical morphological inflection (e.g. the addition of *-ed*, *-ing*, etc.) changes the grammatical category of words, e.g. "These Chinese students at Universities are absolutely mogging the shit out of local born Ricecels."

Affixes provide a combinatorial toolkit for expressing novel concepts, signaling group affiliation, and gaining status within the community. There is evidence in our corpus that illustrates the social process of generating new words. Consider the following example:

*User 1: we should start calling latinos beancels or tacocels. we dont have a proper epiteht for latinocels yet*
*User 2: I've already been doing that. Sushicel, Kimchicel, Tacocel, Schwaramacel, etc*
*User 3: Chollocel, Burritocel, Churrocel, Spiccel*
*User 4: 85 IQ former chicken farmers.*
*User 5: I like spics or beaners*

Once users have learned some of the basic concepts, terminology, and affixes, they can creatively, intentionally generate new words. Participating in word formation is an important part of signaling in-group belonging and an opportunity for recognition by others in the community – the type of acknowledgement and connection that incels are lacking in their offline lives. Moreover, the blends seen in the example above are marked by intention to be shocking or offensive. Taboo words, which reference body parts, body products, sexual acts, ethnic or racial slurs, profanity, vulgarity, or slang have been shown to be more memorable than other emotionally charged words as measured via free recall tasks (Jay 2009, Madan et al. 2017). Taboo words can accomplish positive social goals. Jokes, insider slang, and social commentary all promote group cohesion (Jay 2009, Allan 2018). While these taboo words bring incels together, they are also a defining feature of harassment, discrimination, and verbal abuse.

Blends are exclusive, engaging, and defiant, like the slang teenagers invent or inside jokes. They affirm the belonging and character of the people who come up with them and use them. Blends are not as easy to learn as standalone expressions or new word senses for an already existing word. For example, some religious cults' language is marked by Biblical passages or multiword expressions. These are memorized/lexicalized as one unit for the most part (such as the Twelve Tribes communities' "Anointed One," referring to leader Eugene Spriggs). Similarly, some groups, including incels, will co-opt technical terms from psychology or engineering to add credibility to their ideas, effectively creating a new word sense that is lexicalized like any other (e.g. the word 'valence' from chemistry or math was co-opted by Scientologists to refer to possession by an evil spirit). Blends, on the other hand, generate new affixes that can be combined with other words to make entirely new words. They are lexicalized differently and processed more slowly than non-blends (Johnson et al. 2018), making them excellent tools for shrouding a group's ideas, language, or activity from outsiders who have not devoted the time to learning the ins and outs of the cryptolect.

Overall, the components of incel words tend toward the dehumanizing, racist, and misogynistic, much as the philosophy does. Both authors labeled the terms in Appendix A. Of 64 word components, 46 (71.8%) were dehumanizing, 17 (26.6%) were racist, and 17 (26.6%) were misogynistic. These features reflect a larger theme of incel content, which is that it is designed to be shocking. Users intentionally violate social norms and values to surprise the audience (Dahl et al., 2003). For example, we found one user who coined the term "higlandparkcel", in reference to the Fourth of July Parade shooting in Highland Park, Illinois that killed 7 and injured 48. Celebration of mass shootings, including sharing of live-streamed videos, arguments for the legalization of rape, and extremely violent language are exceptionally common in these forums. This is reflected in the words the community generates as well.

## DISCUSSION

Because the words in this incel lexicon reflect the social mores of the community, the lexicon will support a variety of studies into online radicalization.

We expect there will be patterns where usage increases as men grow more engaged with the forums and the community's ideology. Similarly, decreasing usage of the language may also signal men who are disengaging or de-radicalizing. For example, we selected two highly active users on incels.is and tracked their use of incel language, based on this lexicon, week-by-week over their time in the community. Both users reflect a similar pattern: they joined, used some incel language, took a break, and then rejoined with much higher levels of incel language use (see Figure 1). We do not believe this is the only pattern that will emerge from the data, but it hints at classes of patterns in the use of incel language over time that could be meaningful.

Using the lexicon we present in this paper allows researchers to measure performance of extremism among incels and quantify the response it receives. That would lay the foundation for further research into social proof and influence over genuine beliefs. The incel community provides a type of social support and validation for its members. These men often complain about their social isolation in their offline lives, so a supportive (if strange) online community would fill an important emotional need. However, support from incels only comes to those whose posts are in line with incel ideology. Outsiders, doubters, and questioners are not supported. Furthermore, the more extreme or shocking content is (e.g. celebrating school shootings, a common activity on incel boards), the more support there is for it. This would suggest that men may become radicalized within the incel community because they must perform extremism to receive social support. We hypothesize that after repeatedly saying anti-social things to receive support, over time they come to genuinely believe the extremism.

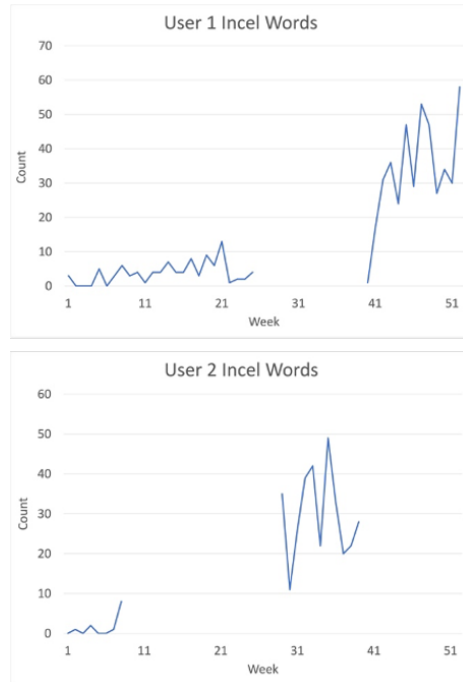

**Figure 1: Incel language usage over time for two users on the forum incels.is.**

To support this hypothesis and better understand the process of radicalization in incel communities, we would need to establish that social support is indeed contingent on extreme behavior. Using the lexicon, we expect future research will address whether there is increased social support for posts that use more radical language. We can measure support by the number of comments on a post and the supportiveness of those posts. "Support" in this case is more challenging to measure than in a typical forum, where sentiment analysis may be effective. Supporting a violent misogynistic post, for example, may come in the form of more misogynistic language. This would be classified a negative sentiment even though it is supportive to the original poster. We expect stance detection (Bestvater & Monroe, 2023; Shugars et al., 2021) would be a more accurate way to measure alignment in stance between the original post and comments as a measure of support. Qualitative analysis of posts and comments in the course of this work may suggest additional heuristics for measuring support.

Using this lexicon, we can also determine a set of patterns of engagement with the language and thus, the incel mindset. Such a typology would, in turn, support a qualitative analysis into common behaviors behind each pattern. For example, we hypothesize that the trends in Figure 1 are pictures of radicalizing men who become increasingly engaged and extreme in their interactions. A close reading of their posts will be critical to understanding what is actually going on to create these patterns.

The lexicon will also support an understanding of the deradicalization process. In future work, we plan to examine language use and social support in the /r/IncelExit community. This is a Reddit-based support group for men who are trying to leave the incel mindset. These men are especially important to this work since they are in the process of disengagement or self-deradicalization *and* they are from the same population of men who participate in the active forums; they were once active users themselves. Quantitatively, we can replicate the radicalized language time series on /r/IncelExit, with a hypothesis that the longer men participate in IncelExit, the less radicalized their language becomes. Replicating the analysis discussed above for increasing radicalization and support, we hypothesize that we will see greater social support for less extreme and more feminist content on /r/IncelExit.

Finally, we also note that while these terms are used almost exclusively in incel communities now, some may creep into broader use over time. The mere use of words in this lexicon should not be considered evidence that someone is necessarily radicalized without additional context. Furthermore, we have shown that there is creativity in the incel community around these words, so the lexicon is likely to continue to evolve. Researchers should be aware that new terms may emerge that they should consider in using this list of words as time passes.

## CONCLUSION

The incel community has an extensive cryptolect, or set of ingroup words that community members can use to identify themselves and that reflect the group's worldview. In this paper, we present an Incel Lexicon that contains the words and building blocks of terminology common to incels. While other researchers have studied incel language, we contribute a lexicon with definitions and affixative elements that can be used both to help new researchers access incel-like content and to drive computational linguistics approaches to understanding radicalization in this community.

We found key features of the incel lexicon to be language which is misogynistic, racist, and dehumanizing, all of which are reflective of incels' worldview. Further, we found that incel language is marked by blends and the affixes they generate, which veil threats of violence and remain difficult for outsiders to understand, yet provide incels with a combinatorial toolkit and the creative linguistic means for effecting shock value, self-affirming their status as a marginalized community, and connecting with each other.